\documentclass[journal]{IEEEtran}

\usepackage{mathtools}
\usepackage{subfigure}
\usepackage{amsmath}
\usepackage{graphicx}
\usepackage{multirow}
\usepackage{makecell}
\usepackage{float}
\usepackage{cite}
\usepackage{indentfirst}
\usepackage{amsthm}
\usepackage{color}
\usepackage{mathrsfs}
\usepackage{amssymb}
\usepackage{amsbsy}
\usepackage{times}  
\usepackage{url}  
\usepackage{verbatim}
\usepackage{hyperref}

\newtheorem*{theorem*}{Theorem}

\newtheorem*{corollary*}{Corollary}

\newtheorem*{lemma*}{Lemma}

\newtheorem*{proposition*}{Proposition}

\ifCLASSINFOpdf

\else

\fi

\begin{document}

\title{Boundary-weighted Domain Adaptive Neural Network for Prostate MR Image Segmentation}

\author{Qikui Zhu, Bo Du*,~\IEEEmembership{Senior Member,~IEEE}, Pingkun Yan*,~\IEEEmembership{Senior Member,~IEEE}%
\thanks{This paper has been accepted by IEEE Trans. Medical Imaging for publication. The published version is available through the \href{https://doi.org/10.1109/TMI.2019.2935018}{DOI of the article}.}%
\thanks{Asterisks indicate co-corresponding authors.}%
\thanks{Q.~Zhu is with School of Computer Science,
Wuhan University, Wuhan, China. (e-mail: QikuiZhu@whu.edu.cn).}%
\thanks{*B.~Du is with School of Computer Science and  State Key Lab of Information Engineering on Survey, Mapping and Remote Sensing, Wuhan University, Wuhan, China. (e-mail: remoteking@whu.edu.cn).}
\thanks{*P.~Yan is with the Department of Biomedical Engineering and the Center for Biotechnology and Interdisciplinary Studies at Rensselaer Polytechnic Institute (RPI), Troy, NY, USA 12180. (e-mail: yanp2@rpi.edu).}%
\thanks{Q.~Zhu and B.~Du were supported by the National Natural Science Foundation of China under Grants 61822113, 41431175, 41871243, and the Natural Science Foundation of Hubei Province under Grants 2018CFA050, 2018CFB432, and the National Key R\&D Program of China under grants 2018YFA0605501 and 2018YFA0605503. P.~Yan was supported by the National Institute of Biomedical Imaging and Bioengineering of the National Institutes of Health under awards R21EB028001 and R01EB027898, and through an NIH Bench-to-Bedside award made possible by the National Cancer Institute. This work was performed while Q.~Zhu was visiting RPI.}%
\thanks{The source code of this work is available at the \href{https://github.com/ahukui/BOWDANet}{GitHub repository}}%
}

\maketitle

\makeatletter
\def\ps@IEEEtitlepagestyle{
	\def\@oddfoot{\mycopyrightnotice}
	\def\@evenfoot{}
}
\def\mycopyrightnotice{
	{\footnotesize
		\begin{minipage}{\textwidth}
			\centering
			Copyright~\copyright~2019 IEEE. Personal use of this material is permitted.  Permission from IEEE must be obtained for all other uses, in any current or future media, including reprinting/republishing this material for advertising or promotional purposes, creating new collective works, for resale or redistribution to servers or lists, or reuse of any copyrighted component of this work in other works.
		\end{minipage}
	}
}

\begin{abstract}
Accurate segmentation of the prostate from magnetic resonance (MR) images provides useful information for prostate cancer diagnosis and treatment. However, automated prostate segmentation from 3D MR images faces several challenges. The lack of clear edge between the prostate and other anatomical structures makes it challenging to accurately extract the boundaries. The complex background texture and large variation in size, shape and intensity distribution of the prostate itself make segmentation even further complicated. Recently, as deep learning, especially convolutional neural networks (CNNs), emerging as the best performed methods for medical image segmentation, the difficulty in obtaining large number of annotated medical images for training CNNs has become much more pronounced than ever. Since large-scale dataset is one of the critical components for the success of deep learning, lack of sufficient training data makes it difficult to fully train complex CNNs. To tackle the above challenges, in this paper, we propose a boundary-weighted domain adaptive neural network (BOWDA-Net). To make the network more sensitive to the boundaries during segmentation, a boundary-weighted segmentation loss is proposed. Furthermore, an advanced boundary-weighted transfer leaning approach is introduced to address the problem of small medical imaging datasets. We evaluate our proposed model on three different MR prostate datasets. The experimental results demonstrate that the proposed model is more sensitive to object boundaries and outperformed other state-of-the-art methods.
\end{abstract}

\begin{IEEEkeywords}
Image segmentation, prostate MR image, domain adaptation, convolutional neural network, boundary-weighted loss.
\end{IEEEkeywords}

\IEEEpeerreviewmaketitle

\section{Introduction}

\IEEEPARstart{A}{ccurately} segmenting prostate magnetic resonance (MR) images plays an important role in prostate diseases diagnosis and treatment, particularly for prostate cancer, which is one of the most common types of cancer in men \cite{pinto2011magnetic}. In clinical practice, medical images can usually be manually segmented by radiologists, which is an expensive and time-consuming process and also prone to inter- and intra-observer variations. Automated segmentation of prostate MR image is highly desirable in clinical practice.
Over the past decade, a number of research groups have proposed various automated prostate segmentation methods. For instance, Shen et~al. \cite{shen2003segmentation} presented a statistical shape model for automatic prostate segmentation in ultrasound images by modeling the shape of the prostate. Guo et~al. \cite{guo2016deformable} proposed a deformable prostate segmentation method, which employed deep feature learning model to extract prostate representation and utilized the sparse patch matching method to infer prostate likelihood map. Tian et~al. \cite{tian2016superpixel} proposed a superpixel-based 3D graph cut algorithm by combining a 3D graph cuts and a 3D active contour model for segmenting the prostate MR images. Although those methods achieved promising performance on prostate segmentation, the complexity of prostate MR images makes it a very challenging problem.

Recently, deep convolutional neural networks (CNNs) have achieved state-of-the-art performance in many fields \cite{huang2017densely,wang2018video,lai2015recurrent,peters2018deep,azizi2018deep, yang2018low, gao2018motion,wang2019ct}, particularly in computer vision and image understanding \cite{simonyan2014very,he2016deep,szegedy2015going}. Many researchers have also employed CNNs in prostate segmentation\cite{Automaticmeyer,tian2018psnet,cheng2016active}. For instance, Milletari et al.~\cite{milletari2016v} proposed a volumetric CNN, which can segment prostate volumes in a fast and end-to-end manner. Yang et al.~\cite{yang2017fine} proposed a novel network, which seamlessly integrates feature extraction, shape prior exploring and boundary estimation together for prostate segmentation. Although great progress has been achieved, there remain challenges that have not been fully addressed, which results in a gap between the clinical needs and the performance of automatic segmentation.

One of the major difficulties in prostate MR image segmentation is that part of the prostate lacks of clear boundary with surrounding tissues, which can be further complicated by complex background texture and large variation in size, shape and intensity distribution of the prostate itself.
Another major challenge is caused by the lack of enough training data, which makes it difficult to get complex networks fully trained as large dataset is a key pillar of the success of CNNs. Thus, the capability of CNNs can be limited for such segmentation tasks.
Facing the above challenges, a number of methods have been proposed from different aspects. For instance, Yu et al.~\cite{yu2017volumetric} designed an efficient volumetric CNN by employing mixed long and short residual connections for improving the training efficiency and discriminating capability under limited training data. Nie et al.~\cite{nie2018asdnet} proposed a region-attention based semi-supervised learning strategy to overcome the challenge that lack of enough training data by employing unlabeled data. To reduce the influence from noise and suppress the tissues around the prostate with similar intensity, Wang et al.~\cite{wang2018deep} developed a novel deep neural network which utilized the attention mechanism to selectively leverage the multi-level features for prostate segmentation.
Although these methods improved the representation capability of network and training efficiency under limited data, obtaining accurate segmentation at slices in the apex and base areas lacking boundary information is still a challenging problem. In addition, efficiently utilizing additional data for training to improve the performance in those difficult locations is yet to be explored.

In this paper, to tackle the above-mentioned challenges and effectively utilize additional datasets to improve network training, a series of experimental settings are designed and tested. A novel boundary-weighted domain adaptive neural network (BOWDA-Net) is then proposed, inspired by the recent progress in adversarial learning \cite{goodfellow2014generative} and transfer learning \cite{ghafoorian2017transfer,luo2017label,zhang2018fully,hoo2016deep,tan2018survey}. The proposed BOWDA-Net employs transfer learning to exploit useful information from other datasets to overcome the challenge of training data shortage. Specifically, to make the process of transfer sensitive to boundaries and to achieve accurate segmentation results even at places with weak boundaries, a boundary-weighted transfer loss (BWTL) is designed to work together with a deep supervision mechanism. Furthermore, to help the image segmentation network quickly converge to segmenting boundaries, we design a boundary-weighted segmentation loss (BWSL) as the supervised loss of segmentation network. Extensive experiments were performed on three prostate image datasets, MICCAI 2012 Prostate MR Image Segmentation (PROMISE12) challenge\footnote{https://promise12.grand-challenge.org/} dataset, Philips 3T MR prostate image dataset, and the Brigham  and  Women's Hospital (BWH) Multiparametric MR (mpMR) prostate image dataset \cite{Data01,fedorov2018annotated}.
The results corroborate the effectiveness of our proposed boundary-weighted domain adaptive neural network (BOWDA-Net). Our method outperformed other state-of-the-art methods and ranked the first in the PROMISE12 challenge.

The remainder of the paper is organized as follows. Section~\ref{sec:related_works} provides a brief review of the related works. Section~\ref{sec:Materials} describes the datasets and Section~\ref{sec:method} presents the proposed BOWDA-Net in detail. In Section~\ref{sec:experiments}, various experiments on prostate MR image segmentation are performed to validate the proposed methods. Finally, several concluding remarks are drawn in Section~\ref{sec:conclusions}.

\section{Related Works}
\label{sec:related_works}

In this section, we briefly review the related works on prostate image segmentation and domain adaptation methods for medical image segmentation.

\subsection{Prostate Image Segmentation}
Accurately segmenting the prostate from images acquired with varying MR protocols and scanners remains a challenge, due to the presence of weak and ambiguous boundaries, as well as the large variability in image contrast and appearance.
Conventional methods tried to deal with these problems using shape priors or image priors like atlases. For example, to increase the robustness of boundary detection for segmenting the prostate in MR images, Gao et al.~\cite{gao2010coupled} proposed a unified shape-based framework to extract the prostate, which consists of two steps, shape registration and shape prior learning. To accurately select atlases for atlas-based image segmentation, Yan~et~at.~\cite{yan2014label} proposed a label image constrained atlas selection method, which exploits the label images to constrain the manifold projection of raw images.

Recently, deep convolutional neural networks (CNNs) have achieved state-of-the-art performances in many fields due to their strong capability in feature representation. CNNs have also been used for prostate image segmentation. Researchers initially used CNNs as feature extractors and then combine them with traditional methods, such as the active shape model or level sets, for feature classification. For example, Cheng et~al.~\cite{cheng2016active} proposed a supervised machine learning model that combines atlas based Active Appearance Model (AAM) with a deep learning model to segment the prostate from MR images. Guo et al.~\cite{guo2015deformable} proposed a deformable segmentation method by unifying deep feature learning with the sparse patch matching. As the use of CNNs in image segmentation advanced, fully convolutional network (FCN) based method was proposed for prostate image segmentation. For instance, Zhu et al.~\cite{zhu2018exploiting} proposed a novel network with bidirectional convolutional recurrent layers to extract both intra-slice and inter-slice information of the prostate for segmentation. Furthermore, to exploit the 3D spatial information, a few studies employed 3D CNNs to extract volumetric features for segmentation. For example, Yu et al.~\cite{yu2017automatic} designed an efficient volumetric CNN by employing mixed long and short residual connections to improve the training efficiency and discriminating capability with limited training data, which outperformed other competitors in MICCAI PROMISE12 challenge in 2017. To robustly and accurately detect the boundary points of the prostate, Brosch~et~al.~\cite{brosch2018deep} formulated
boundary detection as a regression task and employed a convolutional neural network to predict the distances between a surface mesh and the corresponding boundary points, which then achieved the first place of the MICCAI PROMISE12 challenge in 2018.

\begin{figure}
	\centering
	\includegraphics[width=\columnwidth]{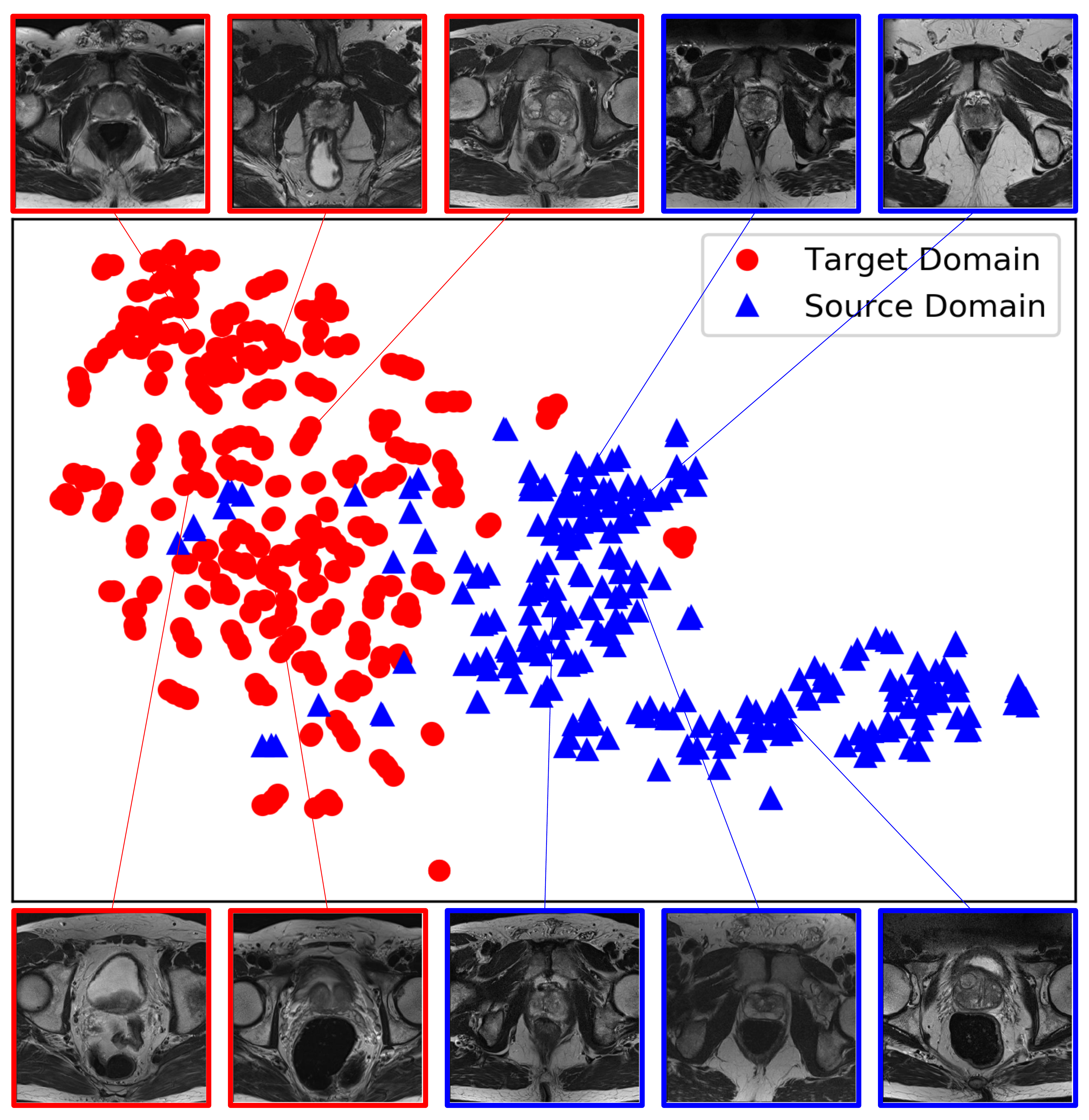}
	\caption{Visualization of the source and target domain images using t-SNE showing the problem of domain shift.}\label{fig:tsne}
\end{figure}

\begin{figure}
	\centering
	\includegraphics[width=\columnwidth]{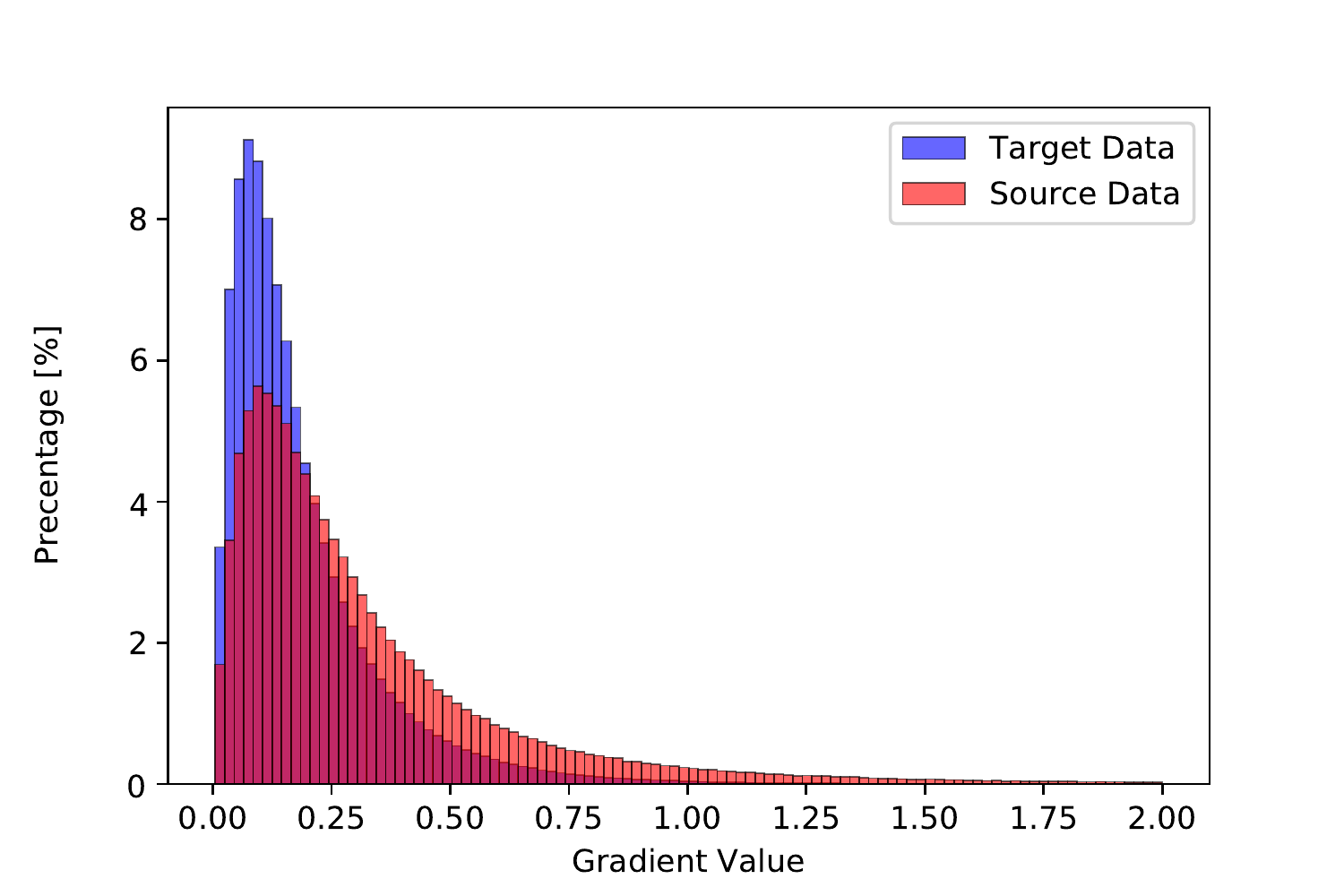}
	\caption{Histograms of the gradient magnitudes at boundary locations from normalized source and target domain datasets, respectively.}\label{fig:edgedis}
\end{figure}

\subsection{Domain Adaptation in Medical Images Segmentation}

Although CNNs have been successfully applied to automated medical image segmentation, such methods suffer from performance degradation when being applied to new datasets different from the training data caused by the problem of domain shift. Recently, several studies have investigated domain adaptation in deep neural networks and applied to medical image analysis tasks. For example, Kamnitsas et al.~\cite{kamnitsas2017unsupervised} developed an unsupervised domain adaptation method for image segmentation by investigating adaptation between databases acquired using two different scanners with difference MR imaging sequences. Ghafoorian et al.~\cite{ghafoorian2017transfer} conducted extensive experiments in white matter hyperintensity segmentation and evaluated the performance of the domain-adapted network with varying sizes of domain data. Goetz et al.~\cite{goetz2015dalsa} tried to employ domain adaptation techniques for effectively correcting the sampling selection errors introduced by the sparse sampling to segment tumor. Recently, Mahmood et al.~\cite{mahmood2018unsupervised} introduced a novel unsupervised reverse domain adaptation framework for addressing the issue of cross-patient network adaptability and limited availability of annotated medical images. The proposed framework first makes real medical images more like synthetic images by employed adversarial training, and meanwhile preserves clinically-relevant features via self-regularization. After that a network trained on a large dataset of synthetically-generated data be applied for
these domain-adapted synthetic-like images. In addition, to overcome the challenge of domain shift between cross-modality medical data, Dou et al.~\cite{dou2018unsupervised} presented a cross-modality domain adaptation framework with unsupervised adversarial learning, which implicitly maps the target data to the feature space of source domain. And Jiang et at.~\cite{jiang2018tumor} presented a tumor-aware, adversarial domain adaptation method for MR image segmentation with unpaired CT and MR images by preserving tumors on synthesized MR images produced from CT image.

\begin{figure*}
	\centering
	\includegraphics[width=\textwidth]{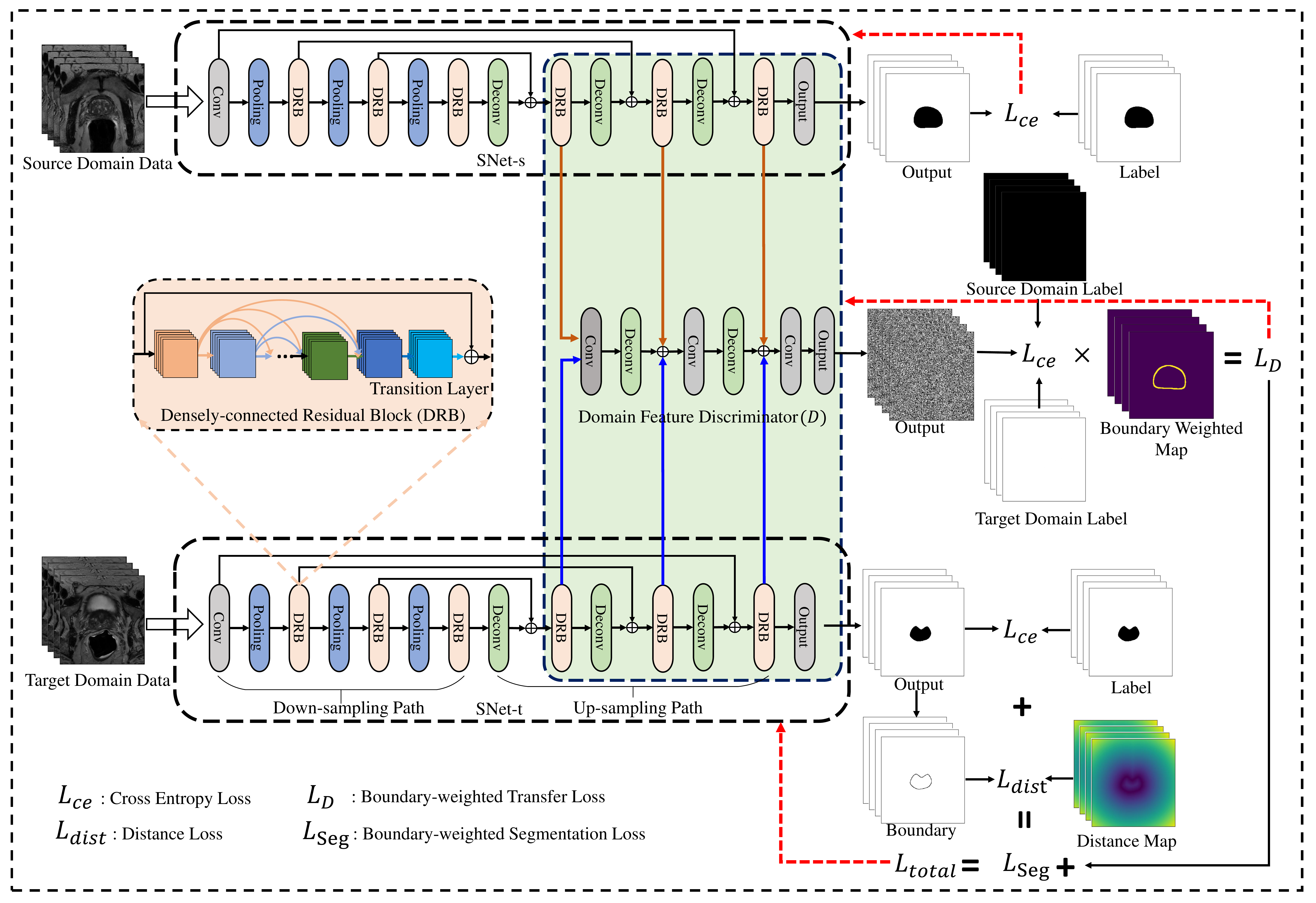}
	\caption{Overview of the proposed boundary-weighted domain adaptive neural network (BOWDA-Net).}
\label{fig:ProposedModel}
\end{figure*}

\section{Materials}
\label{sec:Materials}

In our work, MICCAI 2012 Prostate MR Image Segmentation (PROMISE12) challenge dataset is used as the target domain dataset, a benchmark for evaluating algorithms of segmenting the prostate from MR images. In addition, it is publicly available, performance comparison can be easily performed with other state-of-the-art methods. In that dataset, there are in total 50 transversal T2-weighted MR images of the prostate and the corresponding ground truth segmentation acquired in different hospitals, which were checked and corrected by a radiological resident. These images are a representative set of the types of prostate MR images from multiple vendors and have different acquisition protocols and variations in voxel size, dynamic range, position, field of view and anatomic appearance.

In our experiments, a separate dataset -- 81 prostate MR volumes acquired by a Philips 3T MR scanner with endorectal coil - is used as the source domain dataset. In this dataset, each volume consists of approximately 26 slices and each slice has 512$\times$512 pixels. The in-plane resolution is 0.27mm$\times$0.27mm and the inter-plane distance is 3mm.

To visualize the distribution of the datasets from these two domains, we randomly selected 280 slices from each domain, and then used a pre-trained VGG-16 network~\cite{simonyan2014very} to map each slice to a  feature vector with length of 4096. Then t-SNE~\cite{maaten2008visualizing} was employed to visualize the distribution of the datasets from the two domains as in Fig.~\ref{fig:tsne}, where domain shift between the source and target domains can be observed. Furthermore, to compare the boundary quality of the data from source domain and target domain, we computed the distribution of gradient magnitude at ground truth boundaries over the normalized volumetric images in each dataset with zero mean and unit variance. Fig.~\ref{fig:edgedis} shows the histograms of gradient magnitude at the boundary points from 1324 and 778 images from the source and target domains, respectively. It demonstrates that the boundary quality of data from the source domain is better than that of the data from the target domain.

We also used an additional dataset released by the Brigham and Women's Hospital (BWH) on multi-parametric MR (mp-MR) prostate dataset \cite{Data01,fedorov2018annotated}.
The BWH dataset is composed by the baseline and repeat prostate MR exams for 15 subjects. The scans were obtained with the use of endorectal coil within the period of two weeks.

\section{Boundary-weighted Domain Adaptation}
\label{sec:method}

In this section, we first give an overview of the proposed boundary-weighted domain adaptive neural network (BOWDA-Net) and then present the modules in detail. As shown in Fig.~\ref{fig:ProposedModel}, our proposed BOWDA-Net consists of three main components, which are source domain image segmentation network (SNet-s), target domain image segmentation network (SNet-t) and domain feature discriminator ($D$). During training, SNet-s and SNet-t learn feature representations from source and target domains, respectively. Then the extracted features are delivered to $D$, which is designed to differentiate source domain features from those of target domain. The networks SNet-s, SNet-t and $D$ are designed to work in an adversarial fashion, which is derived from the idea of adversarial learning \cite{goodfellow2014generative,hung2018adversarial}, to overcome the problem of domain shifting and more importantly to exploit the information carried by datasets from source domain to deal with the problems of insufficient training data and weak boundaries in the target domain.

In our experiments, we first train SNet-s with source domain data with cross entropy loss in a supervised manner. The weights of SNet-s are fixed once the training is completed. We then use the obtained weights to initialize SNet-t, which has exactly the same network architecture as SNet-s. After that, the BOWDA-Net is trained in an end-to-end fashion, where SNet-s and SNet-t learn feature representations from source and target domains, respectively, and the discriminator $D$ tries to distinguish the extracted features by their domain. In our proposed BOWDA-Net, the output of $D$ is designed to be in the same size as an input image. Each spatial unit of the output represents the probability of the corresponding input image patch belonging to the target domain. The advantage of such design is to deeply supervise the local patches in the feature map during the process of domain adaptation to differentiate the image details. Furthermore, to make the transfer process focus more on the boundaries and solve the problem of lacking strong edges, in our model, we propose a new boundary-weighted transfer loss (BWTL) for $D$. On top of that, to help the image segmentation network quickly converge to segmenting boundaries, a boundary-weighted segmentation loss (BWSL) is also designed to supervise the training of SNet-t. Details of the proposed method are presented in the following sections.

\subsection{Boundary-weighted Knowledge Transfer}

Transferring information from related data has been shown to be useful in dealing with the problem of lacking sufficient training data \cite{shan2018correction,van2015transfer,hoo2016deep}. However, domain shift caused by the data distribution difference between the datasets is a common problem impacting the efficiency and performance of transfer learning.
Recently, adversarial adaptation methods\cite{hoffman2018cycada,tsai2018learning} have been proposed to deal with the problem, which seek to minimize the between domain distance through minimizing an adversarial loss with respect to a domain discriminator \cite{tzeng2017adversarial,luo2017label}. During training, the representation extractor learns feature representations from source and target domain respectively, the domain discriminator tries to distinguish the features from the source and target domain. When the domain discriminator cannot distinguish the data of source domain from that of target domain, the process of domain adaptation completed and the domain shift problem be addressed. Although existing methods are effective in solving the problem of domain shift and enhancing the performance of transferring learning, the process of transferring is not focused on the information required by the target domain data, which results in the existing method cannot deal effectively with weak boundary.

To   tackle   the   above   mentioned   challenge, in this paper, we propose a supervised boundary-weighted adversarial domain adaptation strategy. In our proposed method, to extract the feature information in source domain, we first train SNet-s under source domain data in a supervised manner, and then freeze the weights. During training, the SNet-s, SNet-t learn feature representations from source and target domain respectively, and then the extracted features be delivered to $D$, which is designed to discriminate source from target domain feature.
However, different from the traditional domain discriminator, in our model, to solve the problem of lacking strong boundary, where the segmentation is the most error-prone, we make the process of information transferring focus more on the boundaries by improving the capability of $D$ in recognizing boundary.
To achieve this goal, we propose a boundary-weighted transfer loss (BWTL) for $D$.
Let ${\{x_s,y_s\}} = \{ (x_s^i,y_s^i)\left| {i = 1,...,m} \right.\}$ represents the training images and ground truths from source domain, and ${\{x_t,y_t\}} = \{ (x_t^i,y_t^i)\left| {i = 1,...,n} \right.\}$ be the training images and ground truths from target domain.
${\mathbf{W}_s}$ and $\mathbf{W}_t$ denote the boundary weighted map of the source and target domain data, which are generated by using the corresponding ground truth labels. The generation process consists of two steps. First, a boundary contour is extracted from the ground truth label using Sobel filter, which is efficiently implemented for GPU computation as part of the training process. Then a $3 \times 3$ Gaussian filter with zero mean and variance ${\sigma ^2} = 0.64$ is employed to filter the boundary map for getting boundary weighted maps. The BWTL for $D$ is defined as
\begin{align}\label{Dloss}
 L_D = & -\mathbb{E}_{x_s} \left[ (1 + \alpha \mathbf{W}_s) \log \left(D \left( \textnormal{SNet-s} (x_s) \right) \right) \right] \\
 & -\mathbb{E}_{x_t} \left[ (1 + \alpha \mathbf{W}_t) \log \left(1 - D \left( \textnormal{SNet-t} (x_t) \right) \right) \right], \nonumber
\end{align}
where $\alpha$ is a weighted coefficient.

\subsection{Boundary-weighted Segmentation Loss}

Generally, for the task of image segmentation, cross entropy $L_{ce}$ is an effective loss function. Let $y$ represents ground truth and $\hat y$ be a segmentation result, $L_{ce}$ can be computed as
\begin{align}\label{celoss}
L_{ce}  =  - \sum\limits_{y} {y\log ( \hat y) + (1 - y)\log (1 - \hat y)}.
\end{align}
However, using cross entropy $L_{ce}$ alone may comprise the segmentation accuracy at boundaries, since the loss may be overwhelmed by the entire region information.
To make the segmentation network more sensitive to the boundaries during segmentation to achieve accurate segmentation, in this paper, a boundary-weighted segmentation loss function (BWSL) is designed. During training, the BWSL utilizes an additional distance loss $L_{dist}$ to regularize the position, shape and continuity of the segmentation to make it close to the object boundaries.
The loss term $L_{dist}$ is defined as
\begin{align}\label{eq_image}
L_{dist} = \beta \sum\limits_{p \in B} {\hat y(p){M_{dist}}(p)},
\end{align}
where  $\hat y$ is a segmentation result, $p$ denotes a point in the point set $B$ containing the boundary points of the segmentation result,  ${M_{dist}}(p)$ is a distance map constructed by the distance transform of the boundary in ground truth label, and $\beta$ is a weighting coefficient.
Accordingly, the BWSL for segmentation network is computed as
\begin{align}\label{eq8}
L_{Seg} = L_{dist}  +  L_{ce}.
\end{align}
In summary, when training SNet-t, a total loss
\begin{align}\label{eq6}
L_{total} = L_{Seg} + {L_D}
\end{align}
will be optimized.

\subsection{Network Design and Configurations}

The details of the networks used in our work are provided in this section. In order to fully leverage the 3D spatial contextual information of volumetric data to accurately segment prostate images, a new 3D network is designed for domain image segmentation network (SNet) with inspiration from the seminal work of U-Net \cite{ronneberger2015u} and DenseNet \cite{huang2017densely}.

As it can be seen in Fig.~\ref{fig:ProposedModel}, SNet-s and SNet-t contain two paths: down-sampling path and up-sampling path. The down-sampling path consists of one convolutional block, three densely-connected residual blocks (DRBs) and three average pooling layers. The pooling layers use stride of two, which gradually reduce the resolution of feature map and increase the receptive field of the convolutional layers. After the down-sampling path, an up-sampling path is attached, which contains three deconvolutional layers and three DRBs. The deconvolutional layers gradually up-sample the feature map until reaching the original size. To further improve the gradient information flow between the down-sampling and up-sampling paths and avoid information loss, inspired by U-Net \cite{ronneberger2015u}, we employ long connections inside the network, which connect the blocks in the same resolution level from the down-sampling and up-sampling paths. Those connections have several advantages. First, they can help effectively propagate context and gradient information both forward and backward between down-sampling and up-sampling paths and alleviate the vanishing-gradient problem. Second, it can help deal with the problem of information loss. To be more specific, when the feature map passes the convolutional and pooling layer, part of the feature information is abandoned and detailed information may be lost. This in turn leads to inaccurate boundaries in the segmentation results. After adding the long connections, the up-sampling path can help retain the feature information from earlier blocks in the down-sampling path to help achieve more accurate segmentation.

The DRB is a new structure proposed in our work  as shown in Fig.~\ref{fig:ProposedModel}, which combines densely connected layers, transition layers, and residual connections together to tackle the problem of overfitting with small training dataset and to promote information propagation within network for faster convergence.
Inside DRB, the densely connections provide direct connections between all subsequent layers and the feature maps produced by all preceding layers are concatenated as input for the subsequent layers. To reduce the number of features and fuse the features from densely connected layers, a transition layer is added at the end of densely connected layers. The transition layer consists of an 1$\times$1 convolutional layer, which reduces the number of feature maps, fuses the feature maps and hence improves the model compactness. To further promote information propagation and make the network easier to optimize, residual connections are employed by DRBs.
Formally, consider an input image ${x_0}$ that is passed through the DRB. Let ${x_l}$ be the output of the $l^{th}$ convolutional layer, $H_l$ is a non-liner transformation of the $l^{th}$ layer and defined as a convolution followed by a batch normalization and a rectifier non-linearity (ReLU). For the DRBs, the output is
\begin{equation}\label{eq4}
{x} = {H_t}({H_l}([{x_0},{x_1},...,{x_{l-1}}])) + {x_0},
\end{equation}
where $[{x_0},{x_1},...,{x_{l-1}}]$ represents the concatenation of the feature maps produced in layers $[0,1,...,l - 1]$, $H_t$ is a non-liner transformation of the transition layer.
Compared with traditional CNNs, the DRBs can easily make the network become deeper and meanwhile possesses fewer parameters, which make the network to be more powerful with hierarchical representation capability.

In summary, the proposed SNet includes convolutional layers, pooling layers, DRBs and deconvolutional layers and has more than 100 layers in depth. The DRBs contain different numbers (4,8,16,8,2) of BN-ReLU-Conv(1$\times$1$\times$1)-BN-ReLU-Conv(3$\times$3$\times$3) with growth rate 32. After each Conv(3$\times$3$\times$3) layer, a dropout layer with 0.3 dropout rate is added to overcome the overfitting problem.
Similar to the referenced works in \cite{long2015learning,luo2017label,dou2018unsupervised} , to make $D$ obtain more useful information and enhance the accuracy of adversarial leaning, in domain discriminator, we take the utilization of multi-level representations into account. The feature representations extracted by each DRB in up-sampling path of SNet-s and SNet-t, total six different features representations, are treated as input of domain discriminator $D$. To eliminate the influence of weight imbalance between supervised loss from SNet-t and adversarial loss from $D$ and make the boundary information be focused, we special design the output of domain discriminator has same size with input and each spatial unit in the output represents the probability of the corresponding image pixel belongs to the target domain.
Inside domain discriminator, we employ three ConvBlocks (Conv(3$\times$3$\times$3)-BN-LeakyReLU) with stride = 1, two deconvolutional layers and one output layer (Conv(1$\times$1$\times$1)) to discriminate source and target domain.

\section{Experiments}
\label{sec:experiments}

\begin{table*}[tp]

  \centering

  \caption{Quantitative evaluation results of BOWDA-Net and other methods on PROMISE12 challenge dataset (by Jan 21, 2019)}
  \label{tab:prostate_comparison}
    \begin{tabular}{l|ccc|ccc|ccc|ccc|c}
    \Xhline{1.2pt}
    \multirow{2}{*}{User}&
    \multicolumn{3}{c|}{ABD [mm]} & \multicolumn{3}{c|}{HD [mm]} & \multicolumn{3}{c|}{DSC [\%]} & \multicolumn{3}{c|}{RVD [\%]} & \multirow{2}{*}{Overall score}\cr\cline{2-13}
    &Whole&Base&Apex&Whole&Base&Apex&Whole&Base&Apex&Whole&Base&Apex&\cr
 \hline

    whu\_mlgroup {\bf(ours)}& 1.35& 1.54 & 1.29 &  4.27 & 4.48 & 3.44  & 91.41 &  89.56 &  89.29 & 4.11 &  1.84 &  3.16 &  {\bf89.59} \cr

    kakatao & 1.29 & 1.47 & 1.40 & 4.14 & 4.32 & 3.77 & 91.76 & 90.05 & 88.27 & 2.11 & 0.39 & 1.89 & 89.54\cr

    sakinis.tomas & 1.34 & 1.51 & 1.44 & 4.15 & 4.41 & 3.79 & 91.33 & 89.73 & 87.95 & 4.63 & 5.83 & 2.54 & 89.44\cr

    pxl\_mcg & 1.40 & 1.59 & 1.40 & 4.28 & 4.35 & 3.56 & 91.23 & 89.08 & 88.55 & 2.08 & -0.07 & 2.23 & 89.39 \cr

    Isensee (nnU-Net)  & 1.31 & 1.45 & 1.46 & 4.00 & 4.05 & 3.79 & 91.61 & 90.29 & 88.05 & 3.42 &1.86 & 3.48 & 89.28 \cr

    segsegseg  &1.37 & 1.51 & 1.44 & 4.38 & 4.36 & 3.67 & 91.37 & 89.85 & 87.60 & 3.06 & 0.38 & 4.12 & 89.13 \cr

    mls.dl.eecs &1.38 & 1.55 & 1.28 & 4.58 & 4.68 & 3.51 & 91.37 & 89.33 & 89.42 & 2.76 & -0.45 & 1.84 & 88.92 \cr

    fly2019 & 1.62 & 1.54 & 1.50 & 5.09 & 4.31 & 3.91 & 90.12 &88.95 & 87.72 & 4.99 &2.19 & 6.65 & 88.73\cr

    rcc & 1.57 & 1.71 & 1.53 & 4.59 & 4.72 & 3.52 & 91.67 & 89.31 & 89.35 & 2.04 & -0.73 & 2.40 & 88.62 \cr

    NPUSAIIP\_JFHealthcare &1.45 & 1.63 & 1.53 & 4.13 & 4.55 & 3.95 & 90.58 & 89.12 &86.89 & 6.68 & 8.60 & -4.49 &88.59\cr
  		\Xhline{1.2pt}

    \end{tabular}
\end{table*}

\subsection{Implementation Details}

In our experiments, due to the variation of PROMISE12 challenge dataset in voxel size, resolution, dynamic range, position, and field of view, we first resampled all the image volumes into a fixed resolution of 0.625mm$\times$0.625mm$\times$1.5mm, and then normalized each volumetric images to have zero mean and unit variance. For the Philips 3T MR image dataset, which has uniform resolution 0.27mm$\times$0.27mm$\times$3mm, we only normalized the intensity of each volumetric images to zero mean and unit variance. The resolution of images in the BWH dataset also varies from 0.27mm$\times$0.27mm$\times$2.9mm to 0.39mm$\times$0.39mm$\times$3.5mm, the in-plane image size is 512$\times$512 pixels. We first resampled each volumetric images into a fixed resolution of 0.27mm$\times$0.27mm$\times$3.0mm, and then normalized the intensities to have zero mean and unit variance. To alleviate the problem of overfitting, data augmentation operations including rotation and flipping are used.
The random cropping strategy is employed to further boost the datasets. During the network training, we randomly cropped sub-volumes in the size of 16$\times$96$\times$96 (${\rm{D}} \times {\rm{W}}\times{\rm{H}}$) voxels from the training data during every iteration.

In the testing phase, the network SNet-t segments sub-volumes of a target image. Similar to other works \cite{yu2017automatic,yu2017volumetric}, we use overlapping sliding windows to crop sub-volumes. In our experiments, the sub-volume size is 16$\times$96$\times$96 (${\rm{D}} \times {\rm{W}} \times {\rm{H}}$) pixels and the stride is 8$\times$48$\times$48 in pixel. The overlapping parts of the output probability maps of these sub-volumes are averaged to get the final volume segmentation.

The proposed method is implemented using the open source deep learning library Keras \cite{keras}. Each model is trained end-to-end with stochastic gradient descent (SGD) optimization method. In the training phase, the learning rate is initially set to 0.0001 and decreased by a weight decay of $1.0 \times 10^{-6}$ after each epoch. The momentum is set to 0.9. The experiments were carried out on a NVIDIA GTX 1080ti GPU with 11GB memory. Due to the limitation of the GPU memory, we chose 4 as the batch size and set the weighted coefficients $\alpha = 1.0$ and $\beta = 0.1$ in Eqns.~(\ref{Dloss}) and (\ref{eq_image}).

\begin{figure}
  \centering
  \includegraphics[width=0.5\textwidth]{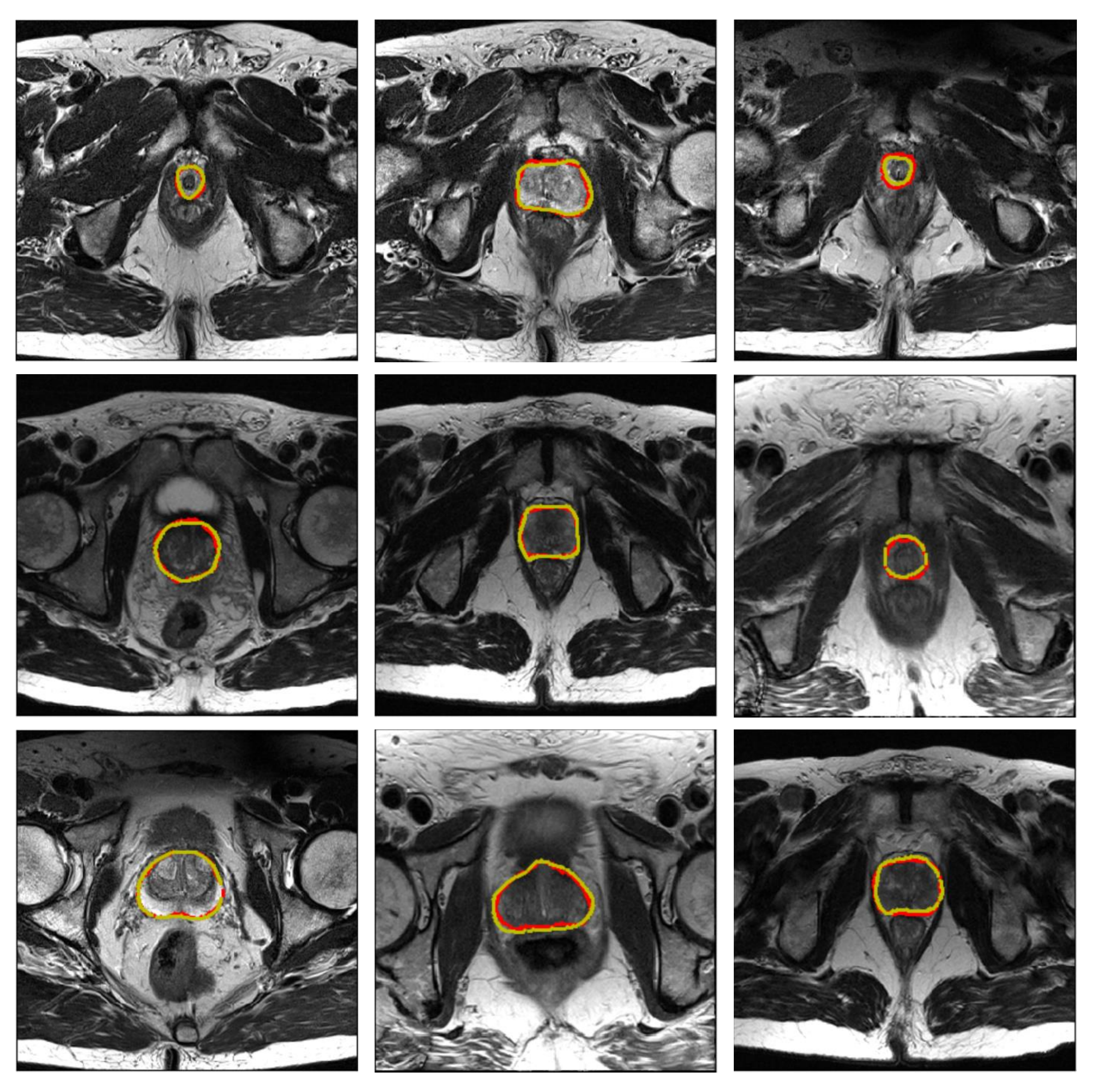}
  \caption{Sample segmentation results of the prostate. The yellow and red contours indicate ground truth and our segmentation results, respectively.}\label{fig:SeRe}
\end{figure}

\subsection{Segmentation Performance}
To evaluate the performance of our BOWDA-Net, we employ PROMISE12 challenge dataset as target domain dataset and Philips 3T MR dataset as source domain dataset in experiment. we compare the results against several other methods, which have also been applied on the MICCAI 2012 Prostate MR Image Segmentation (PROMISE12) challenge dataset. In the PROMISE12 challenge, the organizers provide 30 testing MR images and the corresponding ground truth is held out to evaluate the proposed algorithms. The evaluation metrics used in PROMISE12 challenge include Dice Similarity Coefficient (DSC), the relative volume difference (RVD), average over the shortest distance between the boundary points of the volumes (ABD) and Hausdorff Distance (HD). All the evaluation metrics are calculated in 3D. In addition to evaluating these metrics over the entire prostate segmentation, the challenge organizers also calculated the boundary measures specifically for the apex and base parts of the prostate, because those parts are difficult to segment however very important for many clinical applications. The apex and base are determined by dividing the prostate into three approximately equal sized parts along the axial direction (the first 1/3 as apex and the last 1/3 as base). Then an overall score will be computed by taking all the criteria into consideration rank the algorithms.

The results of our proposed BOWDA-Net and the competitors are shown in Table~\ref{tab:prostate_comparison}. Note that all the results reported in this section were obtained directly from the challenge website\footnote{https://promise12.grand-challenge.org/evaluation/results/} on Jan 21, 2019. Since there are a large number of team submissions, only evaluation scores of the top 10 teams are listed. As it can be seen from Table~\ref{tab:prostate_comparison}, we performed the best and therefore ranked the first place among all the teams with the overall score of 89.59, which demonstrates the advantage of boundary-weighted knowledge transfer and BWSL. Remarkably, the source domain data utilized in BOWDA-Net is not resampled to match the target domain data, which shows that BOWDA-Net can take general similar data to be easily extended to other medical image analysis tasks, especially those with limited training data. Some qualitative results of our method are shown in Fig.~\ref{fig:SeRe}. It is observed that BOWDA-Net can produce accurate segmentation results and delineate the clear contours of prostates in MR images.

\begin{figure*}
  \centering
  \includegraphics[width=0.75\textwidth]{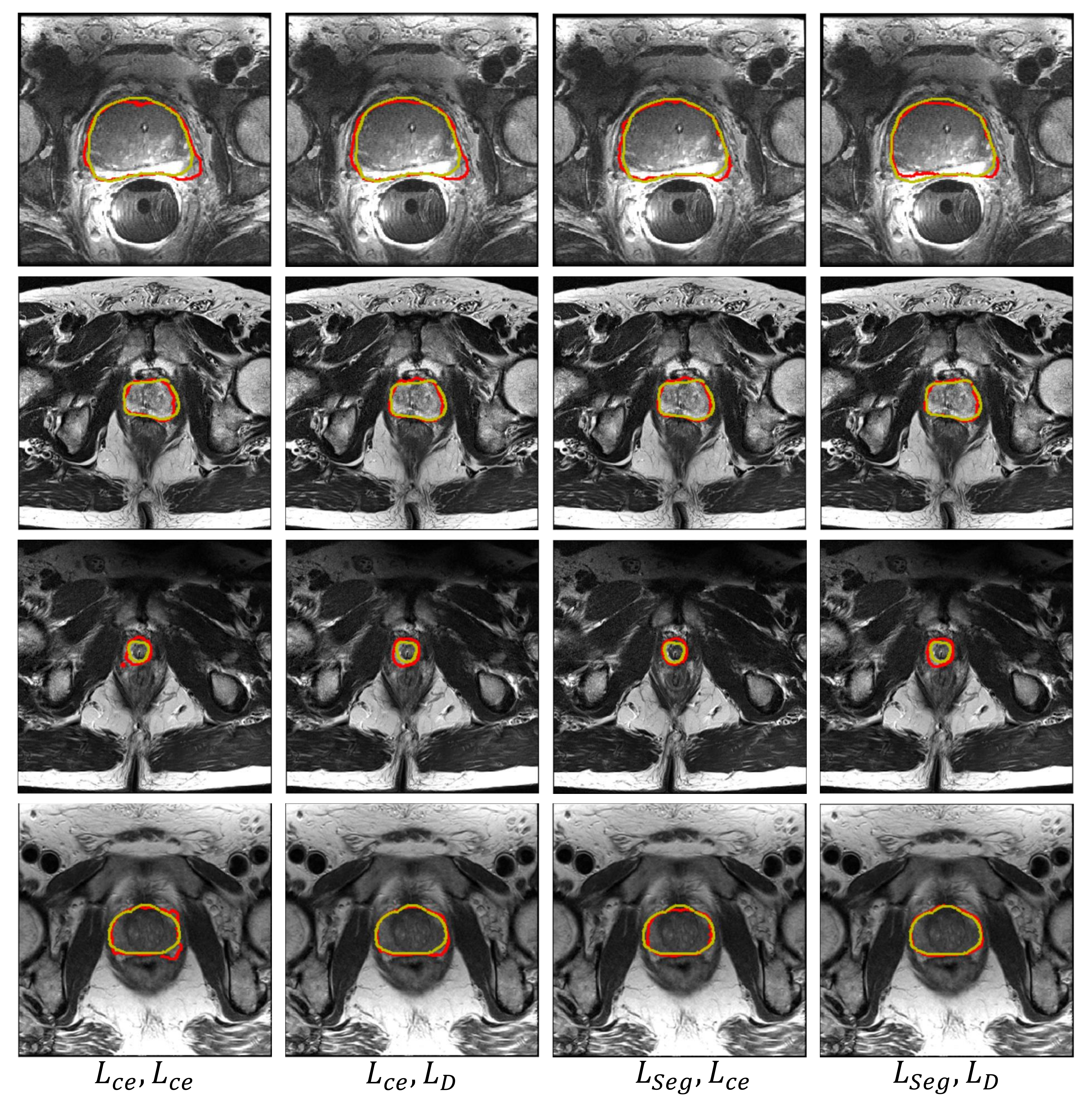}
  \caption{Example segmentation results obtained using different loss functions. The gold standard segmentation is delineated in yellow and the deep learning segmentation results are in red. }\label{fig:DiffLosses}
\end{figure*}

\subsection{Impact of Loss Function}

\begin{table}
	\caption{Effects of loss functions in segmentation performance on PROMISE12 challenge dataset.}
	\centering
	\begin{tabular}{cccccc}
		\Xhline{1.2pt}
		\centering
		SNet-t Loss & ${D}$ Loss & ABD [mm] & HD [mm] & RVD [\%] & DSC [\%]\\
		\hline
		$L_{ce}$ & $L_{ce}$ & \textbf{1.21}  & 11.55  & -3.25  & 90.38 \\
		
		$L_{ce}$ & $L_{D}$ & 2.38  & 12.02  & 4.02  & 90.49\\
		
		$L_{Seg}$ & $L_{ce}$ & 1.65  & 6.83  & 3.71  & 91.47\\
		
		$L_{Seg}$ & $L_{D}$  & 1.58  & \textbf{6.42}  & \textbf{3.24} & \textbf{92.54}\\
		\Xhline{1.2pt}
	\end{tabular}
	\label{Tabel_Loss}
	
\end{table}

To analyze the impact of our proposed BWTL and BWSL on the performance of segmentation, we compared the performance of the proposed BOWDA-Net with different supervised and adversarial losses. Before training, we split the target domain
dataset (PROMISE12 challenge dataset) into two parts by randomly selecting data of 10 subjects for validation and data of the rest 40 subjects for training. The source domain data (Philips 3T MR dataset) employed in this experiment are also not resampled to match the target domain data.
Table~\ref{Tabel_Loss} lists the performances of BOWDA-Net using various combinations of cross entropy loss $L_{ce}$, boundary-weighted transfer loss (BWTL) $L_{D}$ and boundary-weighted segmentation loss (BWSL) $L_{Seg}$.
In addition to employing DSC to evaluate the accuracy of segmentation, we also used ABD, HD, RVD to evaluate the segmentation performance on boundary.

From Table~\ref{Tabel_Loss}, it can be observed that using $L_{Seg}$ and $L_{D}$ as loss functions help achieve better performance than using $L_{ce}$. It demonstrates that BWTL and BWSL can help enhance the performance of networks. In addition, the best performance measured by the majority of the evaluation metrics is achieved when the BWTL and BWSL are both used as loss functions, indicating that the proposed BWTL and BWSL make the trained networks more effective in securing the prostate boundaries. Some segmentation examples from BOWDA-Net with different loss functions are shown in Fig.~\ref{fig:DiffLosses}. It can be seen that the segmentation results produced by BOWDA-Net with BWTL and BWSL have obtained more smoothing and accurate boundaries, which clearly demonstrates that BWTL and BWSL are effective in improving the quality of image segmentation.

\subsection{Effects of Training Strategies}
\label{sec:strategies}

Since all of datasets employed in our experiments are also prostate MR images, mixing two domain datasets together can extend training data directly, which is a basic and straightforward way for solving the problem of lacking training data. On the other hand, fine-tuning a pre-trained network is also a commonly adopted strategy for dealing with this problem, especially when difference exists between the source and target domain datasets. Besides, fine-tuning is also a rudimentary way of transfer learning. In this section, we compare the performances of SNet using different strategies to demonstrate the effectiveness of our proposed BOWDA-Net. The tested training strategies include:
\begin{enumerate}
\item
\emph{Target Domain Training Only}: The target domain data are split into training set $X_{train}^T$ and validation set $X_{val}^T$. Only $X_{train}^T$ is used to train SNet.

\item
\emph{Direct mixing of source and target domains}: We simply mix the source domain data $X^S$ and target domain training set $X_{train}^T$ together to augment the size of training data.

\item
\emph{Mixing after resampling source domain data}: This is similar to the above strategy, except that the source domain data are resampled to the same resolution as the target data.

\item
\emph{Fine-tuning after training in source domain}: We pre-train SNet on source domain data $X^S$ and then fine-tune it on the target data $X_{train}^T$.

\item
\emph{Fine-tuning after training with resampling}: This is similar to the above strategy, except that the source domain data are resampled to the same resolution as the target data.
\item
\emph{The proposed domain adaptation network with cross entropy}: This is similar to the BOWDA-Net, except that the loss function is replaced by cross entropy.
\item
\emph{The proposed domain adaptation strategy (BOWDA-Net).}
\end{enumerate}

\begin{table}[tp]
	\centering
	\caption{Quantitative evaluation of different training strategies on PROMISE12 challenge dataset.}
	\label{tab:Transfer_comparison}

	\begin{tabular}{lclc}
		\Xhline{1.2pt}
		{\bf Strategy} & {\bf DSC [\%]} & {\bf P-Value}\cr\hline
		
		Target domain training only & 88.76 & 0.00988 \cr
		\hline
		
		Direct mixing of source and target domains & 87.78 & 0.01011\cr
		Mixing after resampling source domain data & 89.81 & 0.00070\\
		\hline
		
		Fine-tuning after training in source domain & 89.34 & 0.00192\cr
		Fine-tuning after training with resampling & 89.68 & 0.00038\cr
		\hline
		Domain adaptive network with cross entropy & 90.38 & 0.00852\cr
		\hline		
		BOWDA-Net& {\bf 92.54} & ---------\cr
		
		\Xhline{1.2pt}
	\end{tabular}
\end{table}

In this section, PROMISE12 challenge dataset and Philips 3T MR prostate dataset are employed as target and source domain data, respectively.  Table~\ref{tab:Transfer_comparison} shows the segmentation performances on PROMISE12 challenge dataset using the training strategies described above. It can be seen that directly mixing the source domain data and target domain data has a negative impact on the segmentation performance, which is even worse than using target domain data alone for training. There are two major reasons for that. One of the them is the domain shift problem shown in Fig.~\ref{fig:tsne}. The other one is that the amount of source domain data is larger than the target domain data. Simply mixing the data together would make the network focus more on the source domain rather than the target domain. The SNet then yields poorer performance in the target domain in this case.
This problem is partially remedied by resampling the source domain data to the same resolution as the target domain data, where the DSC value was increased from 87.78\% to 89.81\%.

\begin{figure}
  \centering
  \includegraphics[width=0.5\textwidth]{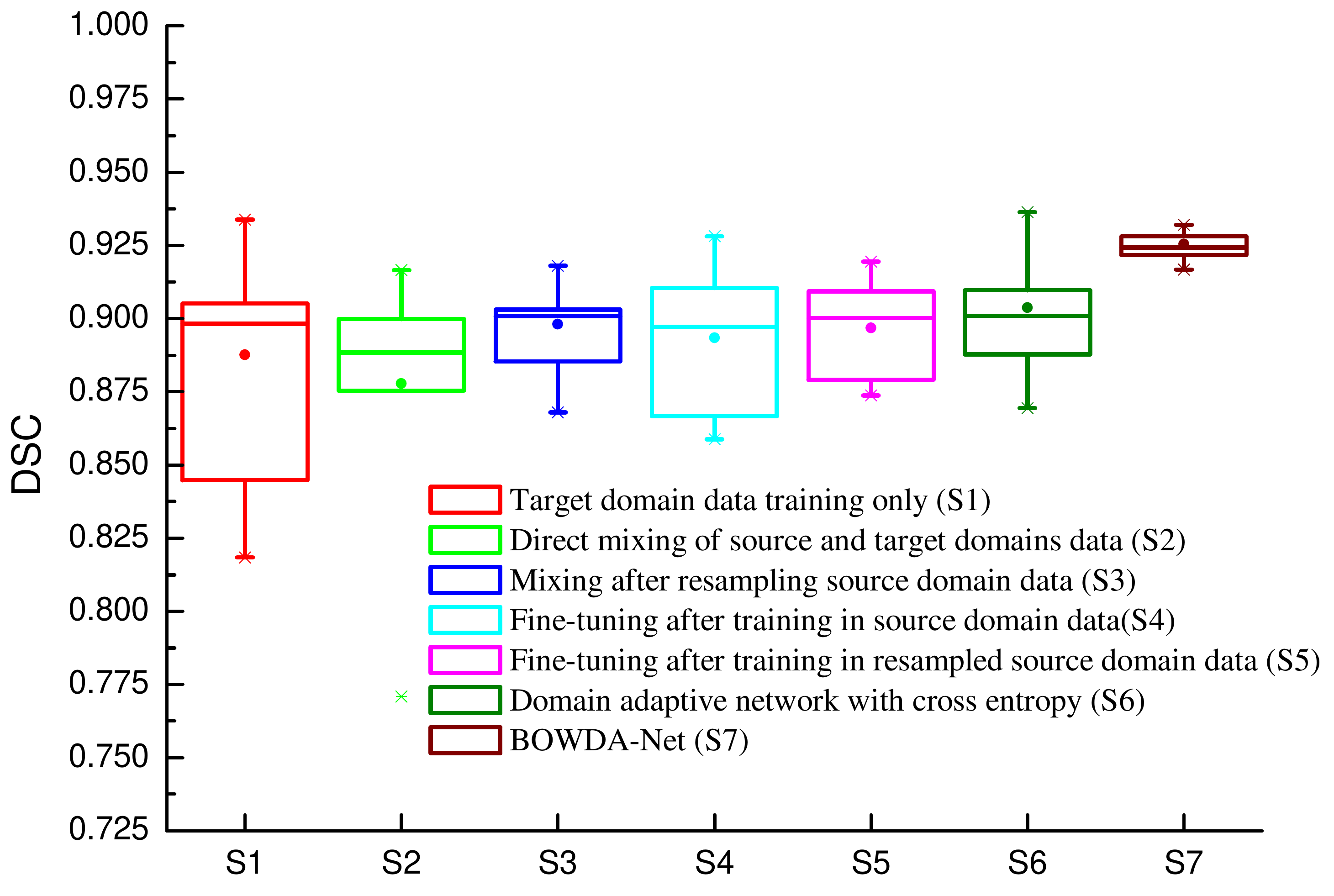}
  \caption{Box plots of the segmentation evaluation results using different training strategies and networks on PROMISE12 challenge dataset.}\label{fig:boxRes}
\end{figure}

Similar effects can be observed on fine-tuning SNet pre-trained in the source domain in Table~\ref{tab:Transfer_comparison}. Compared with pre-training SNet directly on the source domain data, fine-tuning can obtain more accurate segmentation results when the pre-training uses resampled source domain data. However, the performance is still not as good as training SNet by mixing the resampled source domain data and the target domain data. This indicates that the capability of fine-tuning is limited and cannot overcome the problem of lacking sufficient training data.
The influence of the proposed BWTL and BWSL can be observed in Table~\ref{tab:Transfer_comparison}. Compared with the proposed domain adaptive network trained with cross entropy, the network with BWTL and BWSL obtained better results.
Furthermore, we performed statistical comparison of the results using paired $t$-test with a confidence interval of 0.95. BOWDA-Net is compared to the methods for statistical significance, and all the $p$ values are also given in Table~\ref{tab:Transfer_comparison}. It can be seen that the proposed BOWDA-Net significantly outperforms the other methods with $p<0.05$.
In addition, we also summarized the performance of the segmentation methods in Fig.~\ref{fig:boxRes} using box plots. The segmentation results of our proposed BOWDA-Net have much smaller variance and less outliers compared to others.

\subsection{Comparison with Other Transfer Learning Methods}

To demonstrate the effectiveness of our proposed BOWDA-Net through fair comparison using the same 81+50 cases (Philips 3T MR prostate dataset + PROMISE12 challenge dataset), we selected the method submitted by ``chen.junqiang''\cite{chencode} for comparison, because their algorithm is the highest ranked method on the leader board\footnote{https://promise12.grand-challenge.org/evaluation/results/} with published source code\footnote{The source code of our proposed BOWDA-Net can be found at: https://github.com/ahukui/BOWDANet}.
In addition, we also compare our method against the state-of-the-art domain adaptation method by Mehran~et~al.~\cite{javanmardi2018domain} with the same training and test datasets. In their work, two networks, an FCN performing segmentation on the input images called segmentor and a CNN performing classification on the outputs of the segmentor called domain classifier, are combined for image segmentation. The two networks are connected through a gradient reversal layer, which enables adversarial training.

To be fair in evaluating the performance of this domain adaptation method, same like our model, we utilize target domain data label for model training. The model by "chen.junqiang" employs 3D VNet for prostate segmentation, we thus denote the method as VNet thereafter. We also evaluate the performance of Vnet under two training strategies: 1) Directly mixing the source domain data $X^S$ and target domain training set $X_{train}^T$ together to augment the size of training data; 2) Mixing after resampling source domain data, which is similar to the above strategy except that the source domain data are resampled to the same resolution as the target. It is worth noting that the source domain images utilized by BOWDA-Net are also not resampled to match the target domain data. The segmentation performance of the our BOWDA-Net and compared methods are listed in Table~\ref{tab:comparison_with_other}. It can be seen that our proposed BOWDA-Net performed the best and all the performance differences are significant with $p<0.05$.

\begin{table}[tp]
	\centering
	\caption{Quantitative comparison of the proposed method and other existing methods on PROMISE12 challenge dataset.}
	\label{tab:comparison_with_other}
	\begin{tabular}{lclc}
		\Xhline{1.2pt}
		{\bf Method} & {\bf DSC [\%]} & {\bf P-Value}\cr\hline
		
		Mehran et al.~\cite{javanmardi2018domain} & 81.09 & 0.000056 \cr
		\hline
		chen.junqiang\cite{chencode} under strategy-1 & 87.01 & 0.003302\cr\hline
		chen.junqiang\cite{chencode} under strategy-2 & 89.51 &0.001072\\
		\hline		
		BOWDA-Net & {\bf 92.54} & ---------\cr
		
		\Xhline{1.2pt}
	\end{tabular}
\end{table}

\begin{table}[tp]
	\centering
	\caption{Quantitative evaluation of the methods on the BWH dataset.}
	\label{tab:experiment_MRI}
	\begin{tabular}{lclc}
		\Xhline{1.2pt}
		{\bf Strategy} & {\bf DSC [\%]} & {\bf P-Value}\cr\hline
		
		Target domain training only & 85.99 & 0.034 \cr
		\hline
		
		Mixing after resampling source domain data & 87.76 & 0.029\\
		\hline
		
		Fine-tuning after training with resampling & 88.85 & 0.004\cr
		\hline

		BOWDA-Net& {\bf 89.67} & ------\cr
		
		\Xhline{1.2pt}
	\end{tabular}
\end{table}

\subsection{Evaluation on the BWH Dataset}

In this experimental setting, the 81 prostate MR images from Philips 3T MR dataset are also employed as source data. Three subjects from the BWH dataset are randomly selected as test data and the reset are utilized as training data. We then evaluated the performance of our proposed model on this dataset under four different experimental settings as in Section~\ref{sec:strategies}.

Table~\ref{tab:experiment_MRI} shows the segmentation performances using the training strategies described above. Similar to the segmentation results on PROMISE12, our proposed BOWDA-Net gets the highest DSC of 89.67\%. We also compared BOWDA-Net with the rest of the methods for statistical significance. It can be seen that the proposed BOWDA-Net significantly outperforms other methods with $p<0.05$.

\subsection{Network Ablation Study}

\begin{table}
  \caption{Performances of SNet under different ablation configurations on PROMISE12 challenge dataset.}
  \centering
    \begin{tabular}{lc}
		\Xhline{1.2pt}
         Configurations & DSC [\%]\\
        \hline
        FCN & 77.92 \\
        FCN + Dense & 86.02 \\
        FCN + Dense + Residual & 86.94\\
        SNet & \textbf{88.76}\\
   		\Xhline{1.2pt}
    \end{tabular}
    \label{tab:ablation}
\end{table}

In order to evaluate the effectiveness of residual and dense connections in DRBs and long connections used in our proposed SNet, we created four different configurations of our model as follows.
\begin{enumerate}
	\item
\emph{Fully convolutional network (FCN)}: This is indeed  the version of our model without all the dense, residual and long connections.
	\item
\emph{FCN + Dense}: Using only dense connections.
	\item
\emph{FCN + Dense + Residual}: Using both dense and residual connections.
	\item
\emph{The proposed domain image segmentation network (SNet)}:  Using dense, residual and long connections as presented earlier.
\end{enumerate}
Table~\ref{tab:ablation} shows the performance of these networks trained on PROMISE12 challenge dataset by using the target domain data $X_{train}^T$ only. It can be seen that adding residual, long and dense connections can help achieve more accurate segmentation than other networks.

\section{Conclusions}
\label{sec:conclusions}

In this paper, a boundary-weighted domain adaptive neural network (BOWDA-Net) is proposed to address two challenges in prostate image segmentation, which are the lack of clear boundary and the lack of enough annotated data for training CNNs.
Advanced transfer learning method is proposed by incorporating boundary weighting to the scheme.
Extensive experiments on the publicly available PROMISE12 and BWH datasets demonstrate that our proposed method can get more accurate boundaries and achieve superior results compared with other state-of-the-art methods. In our future work, we will extend the proposed method to segment different organs from other imaging modalities.

\bibliographystyle{IEEEtran}
\bibliography{newref}

\begin{thebibliography}{10}
\providecommand{\url}[1]{#1}
\csname url@samestyle\endcsname
\providecommand{\newblock}{\relax}
\providecommand{\bibinfo}[2]{#2}
\providecommand{\BIBentrySTDinterwordspacing}{\spaceskip=0pt\relax}
\providecommand{\BIBentryALTinterwordstretchfactor}{4}
\providecommand{\BIBentryALTinterwordspacing}{\spaceskip=\fontdimen2\font plus
\BIBentryALTinterwordstretchfactor\fontdimen3\font minus
  \fontdimen4\font\relax}
\providecommand{\BIBforeignlanguage}[2]{{%
\expandafter\ifx\csname l@#1\endcsname\relax
\typeout{** WARNING: IEEEtran.bst: No hyphenation pattern has been}%
\typeout{** loaded for the language `#1'. Using the pattern for}%
\typeout{** the default language instead.}%
\else
\language=\csname l@#1\endcsname
\fi
#2}}
\providecommand{\BIBdecl}{\relax}
\BIBdecl

\bibitem{pinto2011magnetic}
P.~A. Pinto, P.~H. Chung, A.~R. Rastinehad, A.~A. Baccala~Jr, J.~Kruecker,
  C.~J. Benjamin, S.~Xu, P.~Yan, S.~Kadoury, C.~Chua \emph{et~al.}, ``Magnetic
  resonance imaging/ultrasound fusion guided prostate biopsy improves cancer
  detection following transrectal ultrasound biopsy and correlates with
  multiparametric magnetic resonance imaging,'' \emph{The Journal of Urology},
  vol. 186, no.~4, pp. 1281--1285, 2011.

\bibitem{shen2003segmentation}
D.~Shen, Y.~Zhan, and C.~Davatzikos, ``Segmentation of prostate boundaries from
  ultrasound images using statistical shape model,'' \emph{IEEE Transactions on
  Medical Imaging}, vol.~22, no.~4, pp. 539--551, 2003.

\bibitem{guo2016deformable}
Y.~Guo, Y.~Gao, and D.~Shen, ``Deformable {MR} prostate segmentation via deep
  feature learning and sparse patch matching,'' \emph{IEEE Transactions on
  Medical Imaging}, vol.~35, no.~4, pp. 1077--1089, 2016.

\bibitem{tian2016superpixel}
Z.~Tian, L.~Liu, Z.~Zhang, and B.~Fei, ``Superpixel-based segmentation for {3D}
  prostate {MR} images,'' \emph{IEEE Transactions on Medical Imaging}, vol.~35,
  no.~3, pp. 791--801, 2016.

\bibitem{huang2017densely}
G.~Huang, Z.~Liu, K.~Q. Weinberger, and L.~van~der Maaten, ``Densely connected
  convolutional networks,'' in \emph{IEEE Conference on Computer Vision and
  Pattern Recognition (CVPR)}, 2017, pp. 4700--4708.

\bibitem{wang2018video}
W.~Wang, J.~Shen, and L.~Shao, ``Video salient object detection via fully
  convolutional networks,'' \emph{IEEE Transactions on Image Processing},
  vol.~27, no.~1, pp. 38--49, 2018.

\bibitem{lai2015recurrent}
S.~Lai, L.~Xu, K.~Liu, and J.~Zhao, ``Recurrent convolutional neural networks
  for text classification.'' in \emph{AAAI Conference on Artificial
  Intelligence (AAAI)}, 2015, pp. 2267--2273.

\bibitem{peters2018deep}
M.~Peters, M.~Neumann, M.~Iyyer, M.~Gardner, C.~Clark, K.~Lee, and
  L.~Zettlemoyer, ``Deep contextualized word representations,'' in
  \emph{Conference of the North American Chapter of the Association for
  Computational Linguistics: Human Language Technologies}, 2018, pp.
  2227--2237.

\bibitem{azizi2018deep}
S.~Azizi, S.~Bayat, P.~Yan, A.~Tahmasebi, J.~T. Kwak, S.~Xu, B.~Turkbey,
  P.~Choyke, P.~Pinto, B.~Wood, P.~Mousavi, and P.~Abolmaesumi, ``Deep
  recurrent neural networks for prostate cancer detection: Analysis of temporal
  enhanced ultrasound,'' \emph{IEEE Transactions on Medical Imaging}, vol.~37,
  no.~12, pp. 2695--2703, 2018.

\bibitem{yang2018low}
Q.~Yang, P.~Yan, Y.~Zhang, H.~Yu, Y.~Shi, X.~Mou, M.~K. Kalra, Y.~Zhang,
  L.~Sun, and G.~Wang, ``Low dose {CT} image denoising using a generative
  adversarial network with wasserstein distance and perceptual loss,''
  \emph{IEEE Transactions on Medical Imaging}, vol.~37, no.~6, pp. 1348--1357,
  2018.

\bibitem{gao2018motion}
Z.~Gao, Y.~Li, Y.~Sun, J.~Yang, H.~Xiong, H.~Zhang, X.~Liu, W.~Wu, D.~Liang,
  and S.~Li, ``Motion tracking of the carotid artery wall from ultrasound image
  sequences: a nonlinear state-space approach,'' \emph{IEEE Transactions on
  Medical Imaging}, vol.~37, no.~1, pp. 273--283, 2018.

\bibitem{wang2019ct}
S.~Wang, K.~He, D.~Nie, S.~Zhou, Y.~Gao, and D.~Shen, ``{CT} male pelvic organ
  segmentation using fully convolutional networks with boundary sensitive
  representation,'' \emph{Medical Image Analysis}, 2019.

\bibitem{simonyan2014very}
K.~Simonyan and A.~Zisserman, ``Very deep convolutional networks for
  large-scale image recognition,'' in \emph{International Conference on
  Learning Representations (ICLR)}, 2015.

\bibitem{he2016deep}
K.~He, X.~Zhang, S.~Ren, and J.~Sun, ``Deep residual learning for image
  recognition,'' in \emph{IEEE Conference on Computer Vision and Pattern
  Recognition (CVPR)}, 2016, pp. 770--778.

\bibitem{szegedy2015going}
C.~Szegedy, W.~Liu, Y.~Jia, P.~Sermanet, S.~Reed, D.~Anguelov, D.~Erhan,
  V.~Vanhoucke, and A.~Rabinovich, ``Going deeper with convolutions,'' in
  \emph{IEEE Conference on Computer Vision and Pattern Recognition (CVPR)},
  2015, pp. 1--9.

\bibitem{Automaticmeyer}
A.~{Meyer}, A.~{Mehrtash}, M.~{Rak}, D.~{Schindele}, M.~{Schostak},
  C.~{Tempany}, T.~{Kapur}, P.~{Abolmaesumi}, A.~{Fedorov}, and C.~{Hansen},
  ``Automatic high resolution segmentation of the prostate from multi-planar
  {MRI},'' in \emph{2018 IEEE 15th International Symposium on Biomedical
  Imaging (ISBI 2018)}, April 2018, pp. 177--181.

\bibitem{tian2018psnet}
Z.~Tian, L.~Liu, Z.~Zhang, and B.~Fei, ``{PSNet}: prostate segmentation on
  {MRI} based on a convolutional neural network,'' \emph{Journal of Medical
  Imaging}, vol.~5, no.~2, p. 021208, 2018.

\bibitem{cheng2016active}
R.~Cheng, H.~R. Roth, L.~Lu, S.~Wang, B.~Turkbey, W.~Gandler, E.~S. McCreedy,
  H.~K. Agarwal, P.~Choyke, R.~M. Summers \emph{et~al.}, ``Active appearance
  model and deep learning for more accurate prostate segmentation on {MRI},''
  in \emph{Medical Imaging 2016: Image Processing}, vol. 9784.\hskip 1em plus
  0.5em minus 0.4em\relax International Society for Optics and Photonics, 2016,
  p. 97842I.

\bibitem{milletari2016v}
F.~Milletari, N.~Navab, and S.-A. Ahmadi, ``V-net: Fully convolutional neural
  networks for volumetric medical image segmentation,'' in \emph{International
  Conference on 3D Vision (3DV)}, 2016, pp. 565--571.

\bibitem{yang2017fine}
X.~Yang, L.~Yu, L.~Wu, Y.~Wang, D.~Ni, J.~Qin, and P.-A. Heng, ``Fine-grained
  recurrent neural networks for automatic prostate segmentation in ultrasound
  images.'' in \emph{AAAI Conference on Artificial Intelligence (AAAI)}, 2017,
  pp. 1633--1639.

\bibitem{yu2017volumetric}
L.~Yu, X.~Yang, H.~Chen, J.~Qin, and P.-A. Heng, ``Volumetric convnets with
  mixed residual connections for automated prostate segmentation from {3D MR}
  images.'' in \emph{AAAI Conference on Artificial Intelligence (AAAI)}, 2017,
  pp. 66--72.

\bibitem{nie2018asdnet}
D.~Nie, Y.~Gao, L.~Wang, and D.~Shen, ``{ASDNet}: Attention based
  semi-supervised deep networks for medical image segmentation,'' in
  \emph{International Conference on Medical Image Computing and
  Computer-Assisted Intervention (MICCAI)}.\hskip 1em plus 0.5em minus
  0.4em\relax Springer, 2018, pp. 370--378.

\bibitem{wang2018deep}
Y.~Wang, Z.~Deng, X.~Hu, L.~Zhu, X.~Yang, X.~Xu, P.-A. Heng, and D.~Ni, ``Deep
  attentional features for prostate segmentation in ultrasound,'' in
  \emph{International Conference on Medical Image Computing and
  Computer-Assisted Intervention (MICCAI)}.\hskip 1em plus 0.5em minus
  0.4em\relax Springer, 2018, pp. 523--530.

\bibitem{goodfellow2014generative}
I.~Goodfellow, J.~Pouget-Abadie, M.~Mirza, B.~Xu, D.~Warde-Farley, S.~Ozair,
  A.~Courville, and Y.~Bengio, ``Generative adversarial nets,'' in
  \emph{International Conference on Neural Information Processing Systems
  (NIPS)}, 2014, pp. 2672--2680.

\bibitem{ghafoorian2017transfer}
M.~Ghafoorian, A.~Mehrtash, T.~Kapur, N.~Karssemeijer, E.~Marchiori,
  M.~Pesteie, C.~R. Guttmann, F.-E. de~Leeuw, C.~M. Tempany, B.~van Ginneken
  \emph{et~al.}, ``Transfer learning for domain adaptation in {MRI}:
  Application in brain lesion segmentation,'' in \emph{International Conference
  on Medical Image Computing and Computer-Assisted Intervention
  (MICCAI)}.\hskip 1em plus 0.5em minus 0.4em\relax Springer, 2017, pp.
  516--524.

\bibitem{luo2017label}
Z.~Luo, Y.~Zou, J.~Hoffman, and L.~F. Fei-Fei, ``Label efficient learning of
  transferable representations acrosss domains and tasks,'' in
  \emph{International Conference on Neural Information Processing Systems
  (NIPS)}, 2017, pp. 165--177.

\bibitem{zhang2018fully}
Y.~Zhang, Z.~Qiu, T.~Yao, D.~Liu, and T.~Mei, ``Fully convolutional adaptation
  networks for semantic segmentation,'' in \emph{IEEE Conference on Computer
  Vision and Pattern Recognition (CVPR)}, 2018, pp. 6810--6818.

\bibitem{hoo2016deep}
H.~Shin, H.~R. Roth, M.~Gao, L.~Lu, Z.~Xu, I.~Nogues, J.~Yao, D.~Mollura, and
  R.~M. Summers, ``Deep convolutional neural networks for computer-aided
  detection: {CNN} architectures, dataset characteristics and transfer
  learning,'' \emph{IEEE Transactions on Medical Imaging}, vol.~35, no.~5, pp.
  1285--1298, 2016.

\bibitem{tan2018survey}
C.~Tan, F.~Sun, T.~Kong, W.~Zhang, C.~Yang, and C.~Liu, ``A survey on deep
  transfer learning,'' in \emph{International Conference on Artificial Neural
  Networks}.\hskip 1em plus 0.5em minus 0.4em\relax Springer, 2018, pp.
  270--279.

\bibitem{Data01}
A.~Fedorov, M.~Schwier, D.~Clunie, C.~Herz, S.~Pieper, R.~Kikinis, C.~Tempany,
  and F.~Fennessy, ``Data from {QIN-PROSTATE-R}epeatability,'' \emph{The Cancer
  Imaging Archive.}, 2018, DOI:10.7937/K9/TCIA.2018.MR1CKGND.

\bibitem{fedorov2018annotated}
------, ``An annotated test-retest collection of prostate multiparametric
  {MRI},'' \emph{Scientific Data}, vol.~5, p. 180281,
  2018,DOI:10.1038/sdata.2018.281.

\bibitem{gao2010coupled}
Y.~Gao, R.~Sandhu, G.~Fichtinger, and A.~R. Tannenbaum, ``A coupled global
  registration and segmentation framework with application to magnetic
  resonance prostate imagery,'' \emph{IEEE Transactions on Medical Imaging},
  vol.~29, no.~10, pp. 1781--1794, 2010.

\bibitem{yan2014label}
P.~Yan, Y.~Cao, Y.~Yuan, B.~Turkbey, and P.~L. Choyke, ``Label image
  constrained multiatlas selection,'' \emph{IEEE Transactions on Cybernetics},
  vol.~45, no.~6, pp. 1158--1168, 2014.

\bibitem{guo2015deformable}
Y.~Guo, Y.~Gao, and D.~Shen, ``Deformable {MR} prostate segmentation via deep
  feature learning and sparse patch matching,'' \emph{IEEE Transactions on
  Medical Imaging}, vol.~35, no.~4, pp. 1077--1089, 2015.

\bibitem{zhu2018exploiting}
Q.~Zhu, B.~Du, B.~Turkbey, P.~Choyke, and P.~Yan, ``Exploiting interslice
  correlation for {MRI} prostate image segmentation, from recursive neural
  networks aspect,'' \emph{Complexity}, vol. 2018, p.~10, 2018.

\bibitem{yu2017automatic}
L.~Yu, J.-Z. Cheng, Q.~Dou, X.~Yang, H.~Chen, J.~Qin, and P.-A. Heng,
  ``Automatic {3D} cardiovascular {MR} segmentation with densely-connected
  volumetric convnets,'' in \emph{International Conference on Medical Image
  Computing and Computer-Assisted Intervention (MICCAI)}.\hskip 1em plus 0.5em
  minus 0.4em\relax Springer, 2017, pp. 287--295.

\bibitem{brosch2018deep}
T.~Brosch, J.~Peters, A.~Groth, T.~Stehle, and J.~Weese, ``Deep learning-based
  boundary detection for model-based segmentation with application to {MR}
  prostate segmentation,'' in \emph{International Conference on Medical Image
  Computing and Computer-Assisted Intervention (MICCAI)}.\hskip 1em plus 0.5em
  minus 0.4em\relax Springer, 2018, pp. 515--522.

\bibitem{kamnitsas2017unsupervised}
K.~Kamnitsas, C.~Baumgartner, C.~Ledig, V.~Newcombe, J.~Simpson, A.~Kane,
  D.~Menon, A.~Nori, A.~Criminisi, D.~Rueckert \emph{et~al.}, ``Unsupervised
  domain adaptation in brain lesion segmentation with adversarial networks,''
  in \emph{International Conference on Information Processing in Medical
  Imaging}.\hskip 1em plus 0.5em minus 0.4em\relax Springer, 2017, pp.
  597--609.

\bibitem{goetz2015dalsa}
M.~Goetz, C.~Weber, F.~Binczyk, J.~Polanska, R.~Tarnawski, B.~Bobek-Billewicz,
  U.~Koethe, J.~Kleesiek, B.~Stieltjes, and K.~H. Maier-Hein, ``{DALSA}: domain
  adaptation for supervised learning from sparsely annotated {MR} images,''
  \emph{IEEE Transactions on Medical Imaging}, vol.~35, no.~1, pp. 184--196,
  2015.

\bibitem{mahmood2018unsupervised}
F.~Mahmood, R.~Chen, and N.~J. Durr, ``Unsupervised reverse domain adaptation
  for synthetic medical images via adversarial training,'' \emph{IEEE
  Transactions on Medical Imaging}, vol.~37, no.~12, pp. 2572--2581, 2018.

\bibitem{dou2018unsupervised}
Q.~Dou, C.~Ouyang, C.~Chen, H.~Chen, and P.-A. Heng, ``Unsupervised
  cross-modality domain adaptation of convnets for biomedical image
  segmentations with adversarial loss,'' in \emph{Proceedings of the 27th
  International Joint Conference on Artificial Intelligence}.\hskip 1em plus
  0.5em minus 0.4em\relax AAAI Press, 2018, pp. 691--697.

\bibitem{jiang2018tumor}
J.~Jiang, Y.-C. Hu, N.~Tyagi, P.~Zhang, A.~Rimner, G.~S. Mageras, J.~O. Deasy,
  and H.~Veeraraghavan, ``Tumor-aware, adversarial domain adaptation from {CT}
  to {MRI} for lung cancer segmentation,'' in \emph{International Conference on
  Medical Image Computing and Computer-Assisted Intervention (MICCAI)}.\hskip
  1em plus 0.5em minus 0.4em\relax Springer, 2018, pp. 777--785.

\bibitem{maaten2008visualizing}
L.~van~der Maaten and G.~Hinton, ``Visualizing data using {t-SNE},''
  \emph{Journal of Machine Learning Research}, vol.~9, no.~11, pp. 2579--2605,
  2008.

\bibitem{hung2018adversarial}
W.-C. Hung, Y.-H. Tsai, Y.-T. Liou, Y.-Y. Lin, and M.-H. Yang, ``Adversarial
  learning for semi-supervised semantic segmentation,'' \emph{The British
  Machine Vision Conference (BMVC)}, 2018.

\bibitem{shan2018correction}
H.~Shan, Y.~Zhang, Q.~Yang, U.~Kruger, M.~K. Kalra, L.~Sun, W.~Cong, and
  G.~Wang, ``{3D} convolutional encoder-decoder network for low-dose {CT} via
  transfer learning from a {2D} trained network,'' \emph{IEEE Transactions on
  Medical Imaging}, vol.~37, no.~12, pp. 2750--2750, 2018.

\bibitem{van2015transfer}
A.~van Opbroek, M.~A. Ikram, M.~W. Vernooij, and M.~de~Bruijne, ``Transfer
  learning improves supervised image segmentation across imaging protocols,''
  \emph{IEEE Transactions on Medical Imaging}, vol.~34, no.~5, pp. 1018--1030,
  2015.

\bibitem{hoffman2018cycada}
J.~Hoffman, E.~Tzeng, T.~Park, J.-Y. Zhu, P.~Isola, K.~Saenko, A.~Efros, and
  T.~Darrell, ``{CyCADA: C}ycle-consistent adversarial domain adaptation,'' in
  \emph{International Conference on Machine Learning (ICML)}, 2018, pp.
  1994--2003.

\bibitem{tsai2018learning}
Y.-H. Tsai, W.-C. Hung, S.~Schulter, K.~Sohn, M.-H. Yang, and M.~Chandraker,
  ``Learning to adapt structured output space for semantic segmentation,'' in
  \emph{Proceedings of the IEEE Conference on Computer Vision and Pattern
  Recognition (CVPR)}, 2018, pp. 7472--7481.

\bibitem{tzeng2017adversarial}
E.~Tzeng, J.~Hoffman, K.~Saenko, and T.~Darrell, ``Adversarial discriminative
  domain adaptation,'' in \emph{IEEE Conference on Computer Vision and Pattern
  Recognition (CVPR)}, 2017, pp. 2962--2971.

\bibitem{ronneberger2015u}
O.~Ronneberger, P.~Fischer, and T.~Brox, ``U-net: Convolutional networks for
  biomedical image segmentation,'' in \emph{International Conference on Medical
  Image Computing and Computer-Assisted Intervention (MICCAI)}.\hskip 1em plus
  0.5em minus 0.4em\relax Springer, 2015, pp. 234--241.

\bibitem{long2015learning}
M.~Long, Y.~Cao, J.~Wang, and M.~I. Jordan, ``Learning transferable features
  with deep adaptation networks,'' in \emph{International Conference on
  International Conference on Machine Learning (ICML)}, 2015, pp. 97--105.

\bibitem{keras}
F.~Chollet \emph{et~al.}, ``Keras,'' \url{https://keras.io}, 2015.

\bibitem{chencode}
J.~Chen, ``{VN}et{3D},'' \url{https://github.com/junqiangchen/VNet3D}, June
  2019.

\bibitem{javanmardi2018domain}
M.~Javanmardi and T.~Tasdizen, ``Domain adaptation for biomedical image
  segmentation using adversarial training,'' in \emph{2018 IEEE 15th
  International Symposium on Biomedical Imaging (ISBI 2018)}.\hskip 1em plus
  0.5em minus 0.4em\relax IEEE, 2018, pp. 554--558.

\end{thebibliography}




\end{document}